\documentclass[10pt,twocolumn,letterpaper]{article}

\usepackage{cvpr}              

\definecolor{cvprblue}{rgb}{0.21,0.49,0.74}
\usepackage[pagebackref,breaklinks,colorlinks,citecolor=cvprblue]{hyperref}
\usepackage[accsupp]{axessibility}
\usepackage{capt-of}
\usepackage{times}
\usepackage{epsfig}
\usepackage{graphicx}
\usepackage{amsmath}
\usepackage{amssymb}
\usepackage{xcolor}
\usepackage{blindtext}
\usepackage{overpic}
\usepackage{transparent}
\usepackage{booktabs}
\usepackage{multirow}
\usepackage[super]{nth}
\usepackage[format=plain,labelformat=simple,labelsep=period,font=small,skip=4pt,compatibility=false]{caption}
\usepackage[font=footnotesize,skip=2pt]{subcaption}
\usepackage{makecell}
\usepackage{enumitem}
\usepackage{algorithm}
\usepackage{algpseudocode}

\makeatletter
\newcommand\notsotiny{\@setfontsize\notsotiny\@vipt\@viipt}

\makeatother

\usepackage{etoolbox}
\usepackage[binary-units]{siunitx}
\robustify\bfseries
\DeclareSIUnit{\inch}{inch}

\newcolumntype{C}[1]{>{\centering\arraybackslash}p{#1}}

\usepackage{xspace}
\makeatletter
\DeclareRobustCommand\onedot{\futurelet\@let@token\@onedot}
\def\@onedot{\ifx\@let@token.\else.\null\fi\xspace}
\def\eg{\emph{e.g}\onedot} 
\def\ie{\emph{i.e}\onedot} 
 
\def\etc{\emph{etc}\onedot}

\makeatother

\usepackage{pifont}
\definecolor{lightgray}{rgb}{0.9, 0.9, 0.9}
\definecolor{lgray}{rgb}{0.66, 0.66, 0.66}
\def\paperID{4429} 
\def\confName{CVPR}
\def\confYear{2024}

\title{FaceChain-ImagineID: Freely Crafting High-Fidelity Diverse Talking Faces from Disentangled Audio}

\author{Chao Xu$^1$$^*$
~ ~ Yang Liu$^1$$^*$
~ ~ Jiazheng Xing$^2$
~ ~ Weida Wang$^2$
~ ~ Mingze Sun$^2$ \\
~ ~ Jun Dan$^2$
~ ~ Tianxin Huang$^3$
~ ~ Siyuan Li$^2$
~ ~ Zhi-Qi Cheng$^4$
~ ~ Ying Tai$^5$
~ ~ Baigui Sun$^1$$^{\dag}$\\
\normalsize $^1$Alibaba Group ~ $^2$FaceChain Community ~ $^3$National University of Singapore \\
\normalsize $^4$Carnegie Mellon University ~ $^5$Nanjing University \\
{\tt\small \{xc264362, ly261666, baigui.sbg\}@alibaba-inc.com}
}

\begin{document}
\twocolumn[{%
\renewcommand\twocolumn[1][]{#1}
\vspace{-2.5em}
\maketitle
\vspace{-3em}
\begin{center}
\centering
\captionsetup{type=figure}
\includegraphics[width=1.0\textwidth]{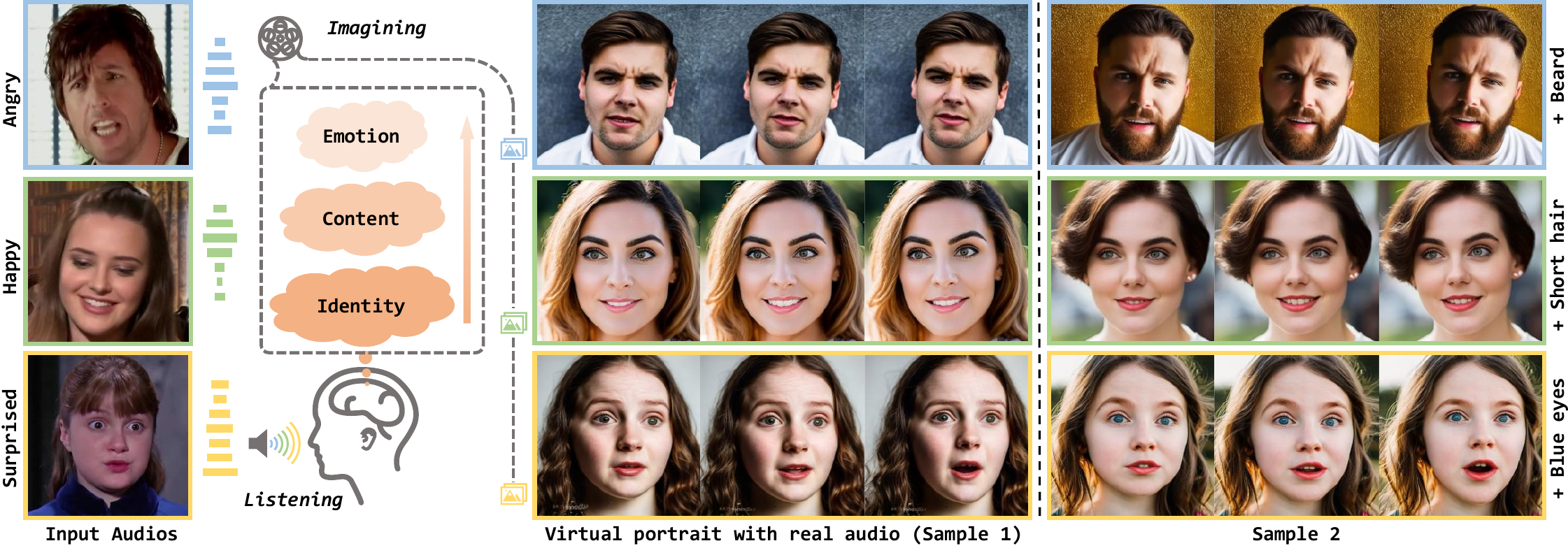}
\caption[figure]{We propose an intuitive new paradigm, Listening and Imagining. \textit{Listening} refers to the progressive disentanglement of identity, content, and emotion from the audio. \textit{Imagining} involves generating diverse videos that align with the given decoupled audio cues while maintaining high consistency within each video. Notably, 
audio-unrelated attributes, \eg, beard, hairstyle, and pupil color, can be freely changed by specified prompts according to personal preferences (Sample 2). Other head pose and eye blink are copied from the reference videos. The audio is the only input to the model, while the left three images are just shown for facial geometry and semantic reference.
\label{fig:teaser}}
\end{center}
}]

\maketitle
\begin{abstract}

\vspace{-0.8em}
In this paper, we abstract the process of people hearing speech, extracting meaningful cues, and creating various dynamically audio-consistent talking faces, termed Listening and Imagining, into the task of high-fidelity diverse talking faces generation from a single audio. Specifically, it involves two critical challenges: one is to effectively decouple identity, content, and emotion from entangled audio, and the other is to maintain intra-video diversity and inter-video consistency. To tackle the issues, we first dig out the intricate relationships among facial factors and simplify the decoupling process, tailoring a Progressive Audio Disentanglement for accurate facial geometry and semantics learning, where each stage incorporates a customized training module responsible for a specific factor. Secondly, to achieve visually diverse and audio-synchronized animation solely from input audio within a single model, we introduce the Controllable Coherent Frame generation, which involves the flexible integration of three trainable adapters with frozen Latent Diffusion Models (LDMs) to focus on maintaining facial geometry and semantics, as well as texture and temporal coherence between frames. In this way, we inherit high-quality diverse generation from LDMs while significantly improving their controllability at a low training cost. Extensive experiments demonstrate the flexibility and effectiveness of our method in handling this paradigm. The codes will be released at \href{https://github.com/modelscope/facechain}{FaceChain}.

\end{abstract}    
\section{Introduction}
\label{sec:intro}

Talking face generation~\cite{zhang2023sadtalker, wang2023seeing, stypulkowski2023diffused, xu2023high, gao2023high, wang2023lipformer, tan2023emmn} is a challenging task that aims to synthesize video based on provided audio and image. This technique finds wide application in various practical scenarios, especially virtual interaction. However, users encountered a dilemma during the process. They are concerned about privacy breaches when using real facial images, while the virtual avatars generated by off-the-shelf methods often fail to align well with their own voices. Thus, we envision a new paradigm if it is possible to liberate the specified source face and directly infer the synchronized \textit{virtual} portrait that matches the \textit{real} audio input. 

Indeed, it is an intuitive process, as people often analyze a voice and then mentally visualize the corresponding video clip. To realize such Listening and Imagining paradigm, there are two critical issues. 1) How to disentangle face-related features \textit{solely} from audio to ensure strong consistency between synthesized videos and given audio. We first explore the natural association between auditory and visual perception. It is true that the facial identity features are closely related to voice characteristics~\cite{bull1983voice, deaton2010understanding, ravanelli2018speaker}. For example, a pronounced chin and prominent brow ridges usually accompany a deep voice, while women and children often have higher pitches. In addition, spoken content involves localized lip movement, and the emotion style reflects the global facial cues~\cite{ekman1978facial, keltner2019emotional}. Accurate decoupling of the above three factors, \ie, identity, content, and emotion, has a significant impact on subsequent generation. However, existing researches~\cite{ji2021audio, peng2023emotalk} mainly focus on the disentanglement of the latter two elements, while other studies~\cite{oh2019speech2face, wen2019face}, although exploring identity, are geared towards generating static face image instead of dynamic animations. 2) How can we maintain diversity between different videos while ensuring consistency within each video by a \textit{single} network. It is the consensus that human imagination is boundless. With the same audio, we can imagine numerous different talking videos, but within each clip, all frames share the uniform information. Benefiting from the progress of diffusion models, we have two potential methods available. One is the combination of text-to-image synthesis, \eg, Latent Diffusion Models (LDMs)~\cite{rombach2022high} and common driving methods like SadTalker~\cite{zhang2023sadtalker} and DiffTalk~\cite{shen2023difftalk}. The other is producing videos by borrowing the framework of text-to-video synthesis~\cite{wu2023tune, esser2023structure, qi2023fatezero, ma2023follow}. However, the former involves two isolated models, while the latter is difficult to perfectly match the control conditions, resulting in noticeable differences between each frame. Additionally, neither of them fully utilizes the audio features.

In this paper, we are dedicated to solving these two problems. Firstly, due to the high degree of coupling in audio, it is challenging to directly employ cross reconstruction under the guidance of pseudo training pairs for disentanglement~\cite{ji2021audio, peng2023emotalk}. To simplify it, we turn to the prior knowledge 3DMM for help and propose a Progressive Audio Disentanglement (PAD) to gradually decouple identity, content, and emotion. As shown in Fig.~\ref{fig:teaser}, we start by predicting the most independent and fundamental identity cues, including the explicit facial geometry and implicit semantics, \ie, shape, gender, and age. Then the estimated shape is used as a condition to disentangle the localized content from the audio. We further supplement the remaining details of global emotion to generate comprehensive facial representations. In this way, we accurately estimate the face-related geometry and semantics from the audio.
For the second concern, we propose a Controllable Coherent Frame (CCF) generation, which offers several appealing designs. Firstly, inspired by the subject-driven models~\cite{ruiz2023dreambooth, shi2023instantbooth}, we combine the implicit audio cues with pred-defined prompt and map them to the CLIP domain~\cite{radford2021learning}, which influence the inferred faces both on semantics and emotions. Secondly, to avoid introducing extra offline components and ensure that the diffusion model has both invariance and variability, we freeze the LDMs and propose three trainable adapters to inject hierarchical conditions, which are equipped with a flexible mechanism that determines whether the outputs have new appearance or remain faithful to the neighboring frame. We further collaborate the CCF with autoregressive inference for diverse and temporally consistent video generation. 

In summary, we present the following contributions.

\begin{itemize}
    \item We propose a new paradigm, Listening and Imagining, for generating diverse and coherent talking faces based on a single audio, which is the first attempt in this field.
   \item We propose a novel PAD that gradually separates the identity, content, and emotion from the audio, providing a solid foundation for accurate and vivid face animations.
   \item We propose a novel frozen LDMs-based CCF with three trainable adapters to faithfully integrate the audio features and address the dilemma of intra-video diversity and inter-video consistency within a single model.
\end{itemize}
\section{Related Work}
\label{sec:related}

\subsection{Audio-driven Talking Face Generation}
Talking face synthesis could be roughly divided into two folds. One of the research directions is GAN-based methods. Early efforts project the audio cues into latent feature space~\cite{chung2017you, zhou2019talking, zhou2021pose, liang2022expressive, wang2023progressive, wang2023seeing, pang2023dpe} and utilize a conditioned image generation framework to synthesize faces. To compensate for information loss in implicit codes, subsequent works have incorporated explicit structural information, such as landmark~\cite{shen2023difftalk, zhong2023identity} and 3DMM~\cite{zhang2023sadtalker, xu2023high, ren2021pirenderer, yin2022styleheat}, to more accurately reflect the audio features on the face. Similarly, 3DMM is used as an intermediate representation in our method. We further leverage it to fully decouple the audio signal. Motivated by progress of the diffusion model in image synthesize tasks, recent approaches~\cite{shen2023difftalk, stypulkowski2023diffused, du2023dae, xu2023multimodal} focus on another research line. They carefully design conditional control modules and train them along with UNet to achieve high-fidelity faces. However, trainable diffusion models fail to inherit many appealing properties, such as diverse generation. 
In this work, we explore the feasibility of using pretrained models to maintain diversity.

\subsection{Audio to Face Generation}
Human voices carry a large amount of personal information, including speaker identity~\cite{deaton2010understanding, ravanelli2018speaker}, age~\cite{grzybowska2016speaker, singh2016relationship}, gender~\cite{li2019improving} and emotion style~\cite{wang2017learning, zhang2019attention}. Based on previous studies, it is possible to directly predict entire face from corresponding audio. Common methods~\cite{oh2019speech2face, wen2019face, bai2022speech, choi2020inference} adopt GANs to generate face images from voice input. However, the above reconstruction-based attempts are not reasonable because audio lacks specific visual features like hairstyles and backgrounds, while these attributes can vary significantly within the same person. Consequently, CMP~\cite{wu2022cross} only models the correlation between facial geometry and voice, but the results lack authenticity. We propose an ideal solution that involves inferring facial structure information from audio and then combining it with a conditional generative model to achieve controllable and diverse generation.

\subsection{Diffusion Models}
Diffusion models~\cite{diffusion1, diffusion2} are popular for generating realistic and diverse samples. In practice, DDIM~\cite{song2020denoising} converts the sampling process to a faster non-Markovian process, while LDMs~\cite{rombach2022high} perform diffusion in a compressed latent space to reduce memory and runtime. Recently, text-to-image generations~\cite{li2023gligen, huang2023composer, saharia2022photorealistic, ramesh2022, ruiz2023hyperdreambooth} have gain significant attentions. For personalized control, DreamBooth~\cite{ruiz2023dreambooth} proposes a subject-driven generation by fine-tuning diffusion models. To reduce training costs, T2I-Adapter~\cite{mou2023t2i} only trains the extra encoder to influence the denoising process. Similarly, our work incorporates the above two characteristics. Another challenging topic is text-to-video generation~\cite{wu2023tune, esser2023structure, zhang2023controlvideo, chen2023control, xing2023simda}. However, they suffer from inconsistency across video frames both temporally and semantically. Additionally, they struggle to generate long videos, impacting the production of high-fidelity talking faces. Therefore, we utilize an autoregressive inference strategy to ensure smooth and consistent transitions between frames, even for long-duration videos.
\section{Method}
\label{sec:method}

\begin{figure*}[t!]
	\centering
	\includegraphics[width=0.95\textwidth]{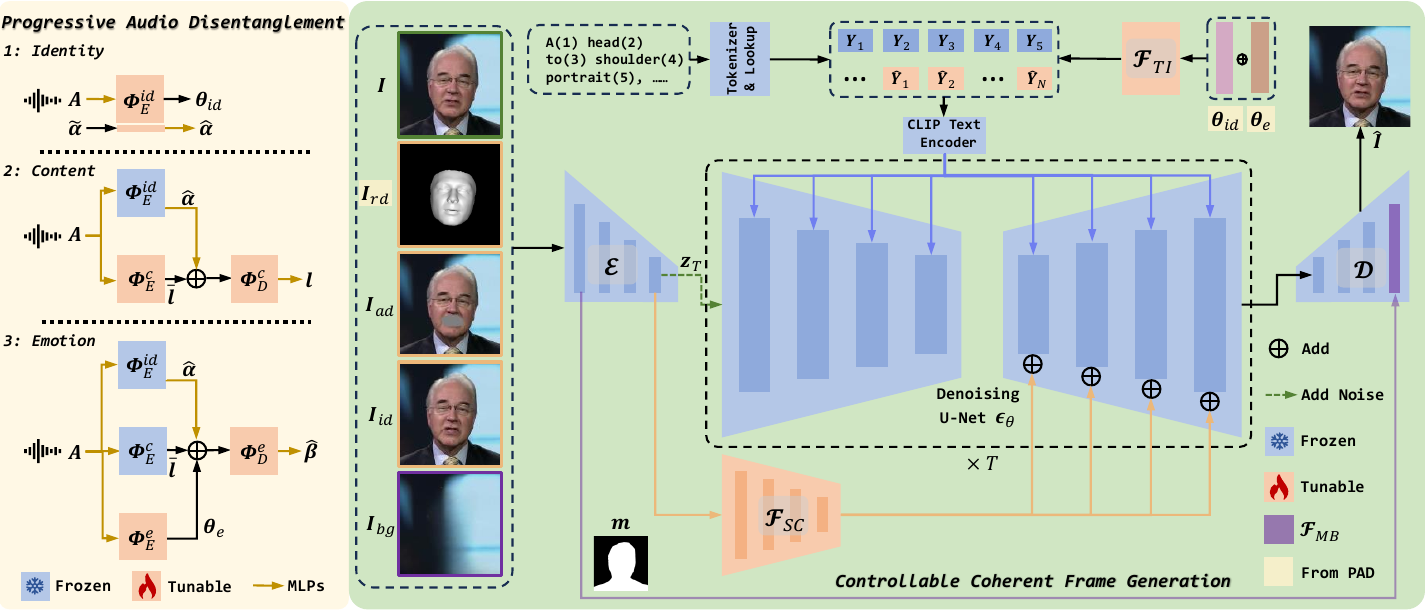}
	\caption{\textbf{Overview of the proposed method.} Our approach involves a two-stage framework that corresponds to the Listening and Imagining. For listening, the Progressive Audio Disentanglement gradually separates the identity, content, and emotion from the entangled audio. For Imagining, Controllable Coherent Frame generation receives the facial semantics ($\boldsymbol{\theta}_{id}$ and $\boldsymbol{\theta}_{e}$ inferred from PAD) and geometry (3D mesh $\boldsymbol{I}_{rd}$ rendered from $\hat{\boldsymbol{\alpha}}$, $\hat{\boldsymbol{\beta}}$ and other coefficients extracted from $\boldsymbol{I}$) to synthesize the diverse audio-synchronized faces, while the $\boldsymbol{I}_{ad}$, $\boldsymbol{I}_{id}$, and $\boldsymbol{I}_{bg}$ are further introduced to achieve highly controllable generation with complete visual and temporal consistency. Please refer to Alg.~\ref{alg:ai} for more details. In this way, we achieve diverse and high-fidelity face animation solely from audio.
	}
	\label{fig:pipeline}
\vspace{-1em}
\end{figure*}

Listening and imagining are instinctive behaviours for humans to perceive and comprehend the world. When people hear a voice, they first analyze the characteristics of the speaker, \ie, who is speaking (Identity), what is being said (Content), and how it feels (Emotion), 
and then imagine diverse and temporally consistent dynamic scenes that match the inferred audio features. To simulate such process, we follow a two-stage phase. Firstly, a Progressive Audio Disentanglement (PAD) is designed to gradually separate the identity, content, and emotion from the coupled audio in Sec.~\ref{sec:3.1}. Secondly, we design Controllable Coherent Frame (CCF) generation (Sec.~\ref{sec:3.2}), which inherits the diverse generation capabilities from Latent Diffusion Models (LDMs) and develops several techniques to enhance controllability: Texture Inversion Adapter (TIA), Spatial Conditional Adapter (SCA), and Mask-guided Blending Adapter (MBA) jointly ensure the generated result meets specific conditions. Autoregressive Inference is utilized to maintain temporal consistency throughout the video. 

\subsection{Progressive Audio Disentanglement}
\label{sec:3.1}
To extract independent cues from the entangled audio signals, previous work~\cite{ji2021audio} builds aligned pseudo pairs and adopts the cross-reconstruction manner for content and emotion disentanglement. 
However, due to the inevitable introduction of errors when constructing pseudo pairs, they become unreliable when more elements are involved. To this end, we only use a single audio for training and adopt 3DMM~\cite{deng2019accurate} to achieve accurate disentangled 3D facial prior, \eg, shape $\boldsymbol{\alpha} \in \mathbb{R}^{80}$, expression sequences $\boldsymbol{\beta} \in \mathbb{R}^{L \times 64}$, \etc, which excludes the face-unrelated factors, such as hairstyle and makeup, and serve as the ground truth for supervision during training. Note that a video shares the same shape while the expression corresponds to each frame. In particular, 
we propose \textit{Progressive Audio Disentanglement} to simplify the decoupling process by extracting the fundamental identity information, then separating the localized mouth movements, and finally obtaining the global expression, outlined in Fig.~\ref{fig:pipeline}. 
For more discussions about the disentanglement order (identity $\rightarrow$ content $\rightarrow$ emotion), the superiority of PAD compared to other methods, and architecture details, please refer to the supplementary materials.

\noindent\textbf{Identity.} 
Motivated by the recent studies that voice is articulatory related to skull shape, as well as gender and age~\cite{maurer1990role, hardcastle2012handbook},
the identity hinted in audio could refer to the facial geometry and semantic embedding. In practice, the identity encoder $\boldsymbol{\Phi}_{E}^{id}$ is the Transformer-based architecture with an extended learnable token $\tilde{\boldsymbol{\alpha}} \in \mathbb{R}^{512}$ of face shape, mapping the MFCC sequence $\boldsymbol{A} \in \mathbb{R}^{L \times 1280}$ to both the estimated shape $\hat{\boldsymbol{\alpha}} \in \mathbb{R}^{80}$ and semantics $\boldsymbol{\theta}_{id} \in \mathbb{R}^{512}$:
\begin{eqnarray}
    &\bar{\boldsymbol{\alpha}}, \bar{\boldsymbol{\theta}}_{id} = \boldsymbol{\Phi}_E^{id}([\tilde{\boldsymbol{\alpha}},\text{MLPs}(\boldsymbol{A})] + \text{PE}), \\
    &\hat{\boldsymbol{\alpha}} = \text{MLPs}(\bar{\boldsymbol{\alpha}}), \boldsymbol{\theta}_{id} = \text{Avg}(\bar{\boldsymbol{\theta}}_{id}),
\end{eqnarray}
where PE is the positional embedding, $\bar{\boldsymbol{\alpha}} \in \mathbb{R}^{512}$, $\bar{\boldsymbol{\theta}}_{id} \in \mathbb{R}^{L \times 512}$. During training, we calculate Reconstruction Loss between predicted $\hat{\boldsymbol{\alpha}}$ and the ground truth $\boldsymbol{\alpha}$: $\mathcal{L}_{id}= \left\|\boldsymbol{\alpha}-\hat{\boldsymbol{\alpha}}\right\|_{2}$. Moreover, we apply Contrastive Loss on semantic embeddings by constructing a positive pair $(\boldsymbol{\theta}_{id}^p, \boldsymbol{\theta}_{id})$ with the same identity and a negative pair $(\boldsymbol{\theta}_{id}^{n}, \boldsymbol{\theta}_{id})$ with the different. Then InfoNCE loss~\cite{oord2018representation} with cosine similarity $\mathcal{S}$ is enhanced between two pairs:
\begin{equation}
    \begin{aligned}
    \mathcal{L}_{con}=-\log \left[\frac{\exp \left(\mathcal{S}\left(\boldsymbol{\theta}_{id}^p, \boldsymbol{\theta}_{id}\right)\right)}{\exp \left(\mathcal{S}\left(\boldsymbol{\theta}_{id}^p, \boldsymbol{\theta}_{id}\right)\right)+\exp \left(\mathcal{S}\left(\boldsymbol{\theta}_{id}^n, \boldsymbol{\theta}_{id}\right)\right)}\right].
    \end{aligned}
    \label{eq:1}
\end{equation}

\noindent\textbf{Content.} The 3DMM expression coefficient involves local lip and global facial movements, while the spoken content is mainly related to the former. Inspired by the work~\cite{wang2023seeing}, we employ the lip reading expert~\cite{shi2022learning} into the training phase and produce the only mouth-related expression coefficients $\boldsymbol{l} \in \mathbb{R}^{L \times 64}$ for content disentanglement. As shown in Fig.~\ref{fig:pipeline}, we construct a Transformer-based encoder-decoder architecture, where the linguistic features $\bar{\boldsymbol{l}} \in \mathbb{R}^{L \times 64}$ embedded by the content encoder $\boldsymbol{\Phi}_{E}^{c}$ and the identity template output by the frozen identity encoder are combined together and sent into the content decoder $\boldsymbol{\Phi}_{D}^{c}$ to obtain $\boldsymbol{l}$:
\begin{eqnarray}
    &\bar{\boldsymbol{l}} = \boldsymbol{\Phi}_E^{c}(\text{MLPs}(\boldsymbol{A}) + \text{PE}), \\
    &\boldsymbol{l} =  \text{MLPs}(\boldsymbol{\Phi}_D^{c}(\bar{\boldsymbol{l}} + \text{MLPs}(\hat{\boldsymbol{\alpha}}))).
\end{eqnarray}
We train this stage by using two losses: Regularization Loss calculates the distance between $\boldsymbol{\beta}$ and $\boldsymbol{l}$ with a relative small weight to smooth training phase: $\mathcal{L}_{reg}= \left\|\boldsymbol{l}-\boldsymbol{\beta}\right\|_{2}$. Besides, we project the coefficients to the image domain with textures, and utilize lip-reading expert to predict the text $\hat{\boldsymbol{X}}$. Assuming that the original text content is $\boldsymbol{X}$, we compute the Lip-reading Loss via cross-entropy: $\mathcal{L}_{lip}= - \boldsymbol{X} \text{log} P (\hat{\boldsymbol{X}} | \boldsymbol{V})$, where $\boldsymbol{V}$ is the ground truth video.

\noindent\textbf{Emotion.} In this stage, we directly utilize $\boldsymbol{\beta}$ as constraint to decouple the remaining emotion styles. As shown in Fig.~\ref{fig:pipeline}, based on the second stage, an additional emotion encoder $\boldsymbol{\Phi}_{E}^{e}$ is introduced to generate the pooled emotion embeddings $\boldsymbol{\theta}_{e} \in \mathbb{R}^{512}$, which is combined with identity $\hat{\boldsymbol{\alpha}}$ and linguistic features $\bar{\boldsymbol{l}}$ for the expression coefficient prediction by the emotion decoder $\boldsymbol{\Phi}_{D}^{e}$:
\begin{eqnarray}
    &\boldsymbol{\theta}_{e} = \text{Avg}(\boldsymbol{\Phi}_E^{e}(\text{MLPs}(\boldsymbol{A}) + \text{PE})), \\
    &\hat{\boldsymbol{\beta}} =  \text{MLPs}(\boldsymbol{\Phi}_D^{e}(\boldsymbol{\theta}_{e} + \bar{\boldsymbol{l}} + \text{MLPs}(\hat{\boldsymbol{\alpha}}))).
\end{eqnarray}
Similar to Eq.~\ref{eq:1}, we adopt Contrastive Loss to extract more discriminative emotion features, details of which are omitted here. Additionally, the regular Reconstruction Loss is used to supervise generated $\hat{\boldsymbol{\beta}}$.

\subsection{Controllable Coherent Frame Generation}
\label{sec:3.2}
Existing methods typically specify the input source face along with audio cues for video generation, while our objective is to achieve visually diverse and audio-consistent animation directly from input audio. Recent LDMs are naturally suited to diverse generation, and their conditional text prompts offer a pathway for freely editing attributes that cannot be deduced from the audio, as shown in Fig.~\ref{fig:teaser}. In exchange, they are relatively weak in controllability, especially in video generation~\cite{wu2023tune, qi2023fatezero}. Thus, to tailor it to our task without introducing extra offline models (\eg, LDMs + DiffTalk\cite{shen2023difftalk}), we must tackle two challenges: ensuring that the synthesized video content aligns with the given conditions, and achieving smooth temporal transitions across frames. Our targeted designs are depicted in Fig.~\ref{fig:pipeline}.

\noindent\textbf{Textual Inversion Adapter.}
The critical issue in injecting the identity $\boldsymbol{\theta}_{id}$ and emotion $\boldsymbol{\theta}_{e}$ inferred from PAD to the frozen LDMs lies in aligning them with the CLIP domain. Inspired by the inversion technique~\cite{ruiz2023dreambooth}, we propose a \textit{Textual Inversion Adapter} $\boldsymbol{\mathcal{F}}_{TI}$, which maps the input vector-based conditions into a set of token embeddings to sufficiently represent these conditions in the CLIP token embedding space.
In detail, we first predefine the prompt for basic high-quality face generation. It is tokenized and mapped into the token embedding space by employing a CLIP embedding lookup module, obtaining $\left\{\boldsymbol{Y}_{1},\dots,\boldsymbol{Y}_{M}\right\}$. Then, the adapter encodes $\boldsymbol{\theta}_{id}$ and $\boldsymbol{\theta}_{e}$ into the aligned pseudo-word token embeddings $\left\{\hat{\boldsymbol{Y}}_{1},\dots,\hat{\boldsymbol{Y}}_{N}\right\}$. We concatenate these two embeddings and feed them to the CLIP text encoder, whose output $\boldsymbol{Y}$ is applied on the cross-attention layer of LDMs to guide the denoising process. When training this adapter, we freeze all of the other parameters:
\begin{equation}
  \begin{aligned}
    \mathbb{E}_{\boldsymbol{z}_0, \boldsymbol{\varepsilon} \sim N(0, \boldsymbol{I}), t, \boldsymbol{Y}}\left\|\boldsymbol{\varepsilon}-\boldsymbol{\varepsilon}_\theta\left(\boldsymbol{z}_t, t, \boldsymbol{Y}\right)\right\|_2^2.
  \end{aligned}
\end{equation}

\noindent\textbf{Spatial Conditional Adapter.} Inspired by T2I-Adapter, we devise \textit{Spatial Conditional Adapter} $\boldsymbol{\mathcal{F}}_{SC}$ to further fuse the explicit conditions. As shown in Fig.~\ref{fig:pipeline}, 3D face mesh $\boldsymbol{I}_{rd}$ 
contains rich audio-synchronized facial geometry, \ie, face shape, lip movement, and expression style. Besides, we sample a random frame $\boldsymbol{I}_{id}$ of the same identity to provide the face appearance and background. These two conditions are enough for common methods~\cite{zhang2023sadtalker, zhou2021pose} to animate the identity face to desired pose and expression. However, it is difficult for frozen LDMs to learn complex spatial transformation. Thus, we further incorporate the adjacent frame $\boldsymbol{I}_{ad}$ as an additional reference, and mask its mouth area $\boldsymbol{\mathcal{M}}$ produced by 3D mesh to avoid the networks from learning shortcuts by directly copying from it. This condition makes the deformation learning much easier and also provides motion cues.
We only train this adapter in this stage:
\begin{equation}
  \begin{aligned}
    \mathbb{E}_{\boldsymbol{z}_0, \boldsymbol{\varepsilon} \sim N(0, \boldsymbol{I}), t, \boldsymbol{F}_{sc}}\left\|\boldsymbol{\varepsilon}-\boldsymbol{\varepsilon}_\theta\left(\boldsymbol{z}_t, t, \boldsymbol{Y}, \boldsymbol{F}_{sc}\right)\right\|_2^2,
  \end{aligned}
\end{equation}
where $\boldsymbol{F}_{sc}=\boldsymbol{\mathcal{F}}_{SC}(\boldsymbol{\mathcal{E}}(\boldsymbol{I}_{rd})$ + $\boldsymbol{\mathcal{E}}(\boldsymbol{I}_{id})$ + $\boldsymbol{\mathcal{E}}(\boldsymbol{I}_{ad}))$ when $p \textless 0.8$, else $\boldsymbol{\mathcal{F}}_{SC}(\boldsymbol{\mathcal{E}}(\boldsymbol{I}_{rd}))$. $\boldsymbol{\mathcal{E}}$ is the VAE encoder. $p$ is a random number generator within the range from 0 to 1.

\begin{algorithm}[t!]
\caption{Autoregressive Inference}
\label{alg:ai}
\begin{algorithmic}

\small 

\State \textbf{Input:} $\boldsymbol{z}_T$:  \text{random noise} \\
$\left\{\boldsymbol{I}_{rd}^{i}\right\}_{i=1}^H$: \text{3D mesh sequences}   $\triangleright$ Inferred from audio\\
$\boldsymbol{Y}$: \text{token embeddings}   $\triangleright$ Inferred from audio\\
\\
$\triangleright$ $i=1$, Diverse mode \\
$\hat{\boldsymbol{I}}_{1}=\boldsymbol{\mathcal{D}}(\text{CCF}(\boldsymbol{z}_T, \boldsymbol{\mathcal{E}}(\boldsymbol{I}_{rd}^{1}), \boldsymbol{Y})$
\\
\\
$\triangleright$ $i=2$...$H$, Coherent mode 
\For{$i = 2...H$}
    \State $\boldsymbol{I}_{id}^{i}=\hat{\boldsymbol{I}}_1$, $\boldsymbol{I}_{bg}^{i}=\boldsymbol{\mathcal{G}}_{IA}(\hat{\boldsymbol{I}}_1)$
    \State $\boldsymbol{I}_{ad}^{i}=\boldsymbol{\mathcal{M}}(\hat{\boldsymbol{I}}_{i-1})$,
    $\boldsymbol{m}_{i}=\boldsymbol{\mathcal{G}}_{MOD}(\hat{\boldsymbol{I}}_{i-1})$
    \State $\hat{\boldsymbol{I}}_{i}=\boldsymbol{\mathcal{D}}^{'}(\text{CCF}(\boldsymbol{z}_T, \boldsymbol{\mathcal{E}}([\boldsymbol{I}_{rd}^{i}, \boldsymbol{I}_{id}^{i}, \boldsymbol{I}_{ad}^{i}]), \boldsymbol{Y}), \boldsymbol{\mathcal{E}}(\boldsymbol{I}_{bg}^{i}), \boldsymbol{m}_{i})$

\EndFor
\State {\bf Output:} $\hat{\boldsymbol{V}}$: generated video 

\end{algorithmic}
\end{algorithm}

\begin{figure}[t!]
	\centering
	\includegraphics[width=0.48\textwidth]{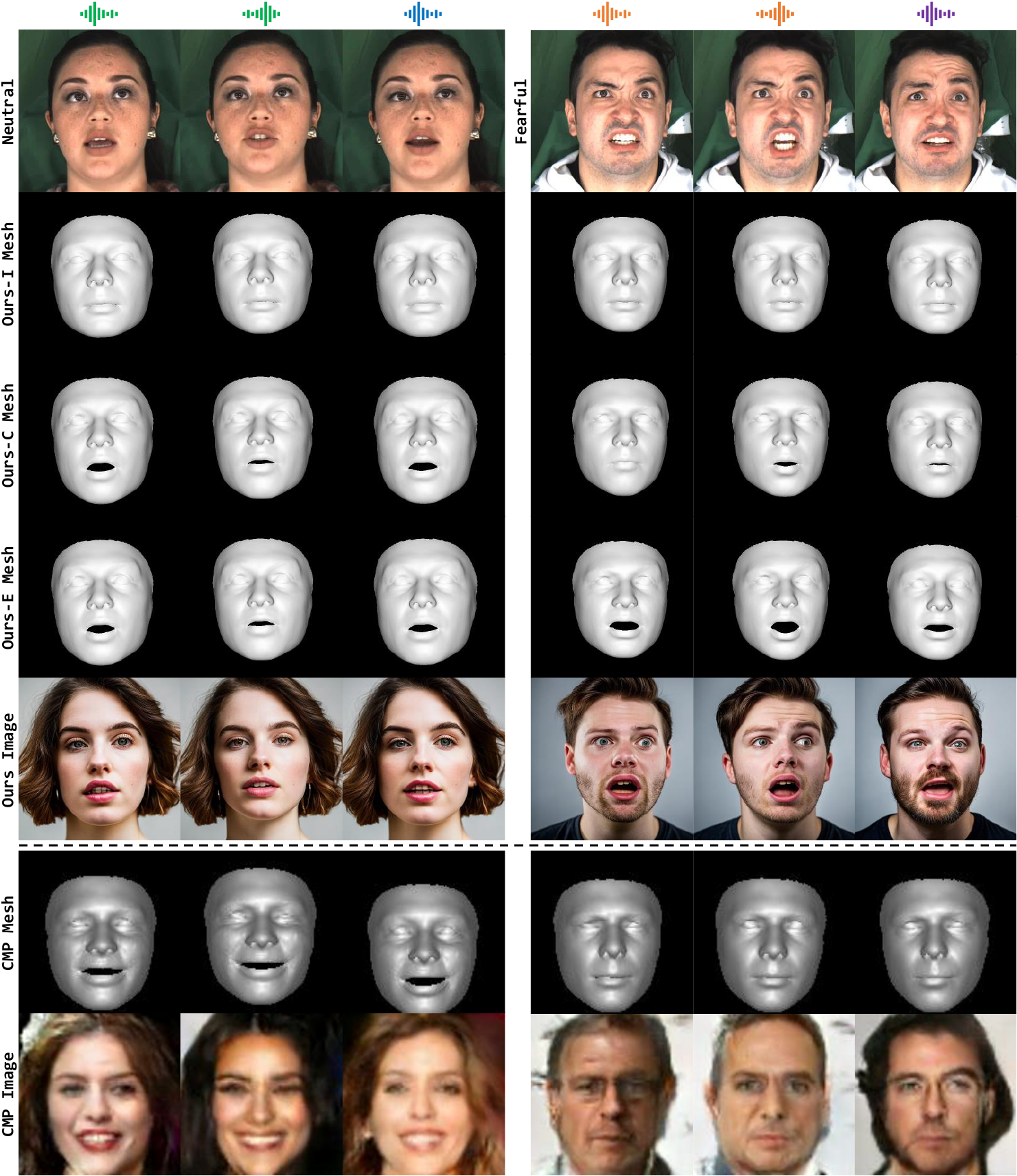}
	\caption{Qualitative results of audio-to-face on MEAD . Icons of the same color indicate samples from the same audio. Ours-I, -C, and -E mean the stage of identity, content, emotion decoupling.
	}
        \vspace{-1.5em}
	\label{fig:exp1_1}
\end{figure}

\begin{figure*}[t!]
	\centering
	\includegraphics[width=0.88\textwidth]{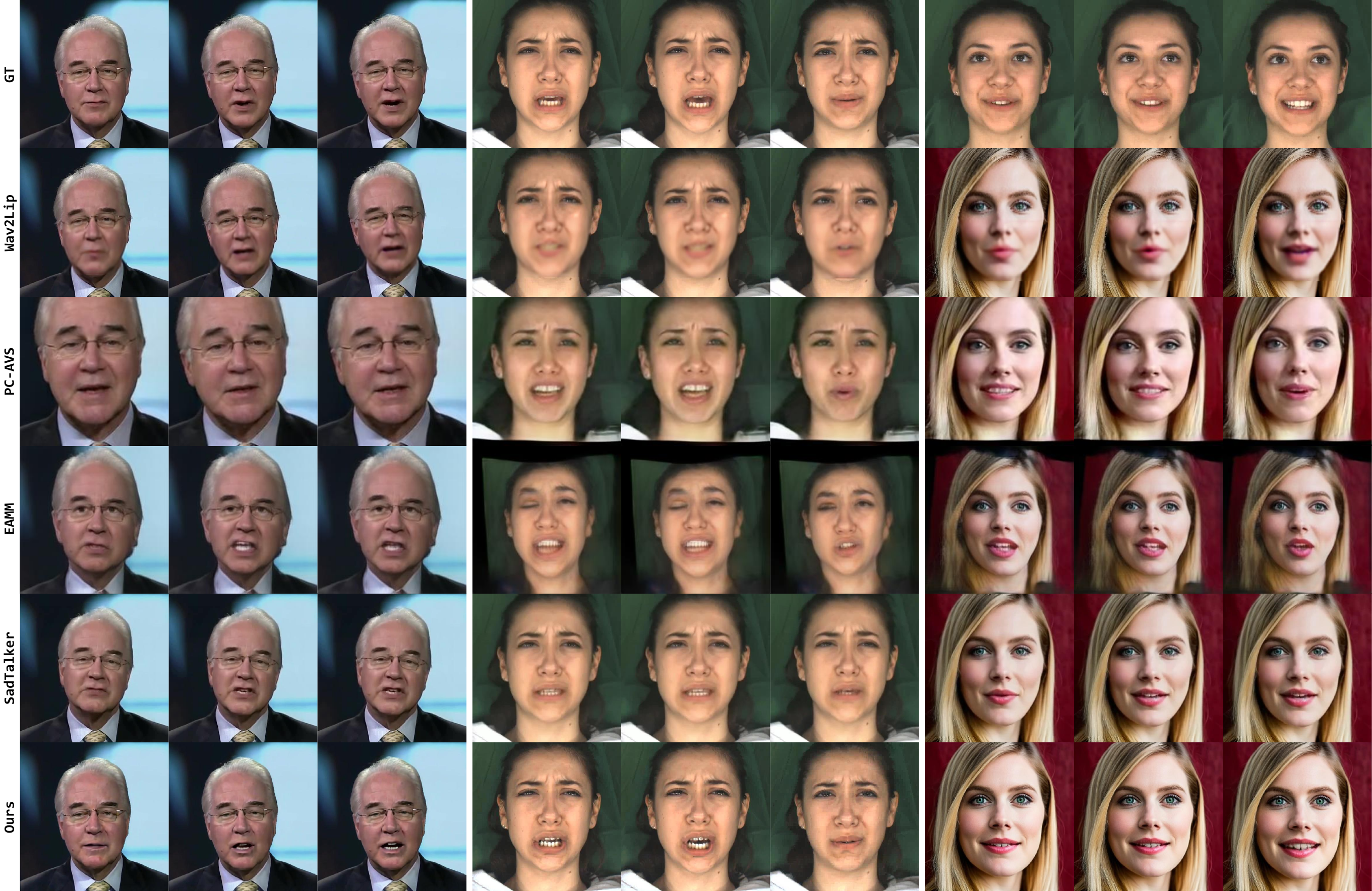}
	\caption{Visual comparison with recent SOTA methods. Images are from officially released codes for fair comparisons. The first sample selected from HDTF, second from MEAD. For the third, based on the audio of the first row, our method generate a unseen face for all competitors, which using this as the source face to produce talking faces. The first row provides the ground truth for facial expression.
	}
	\label{fig:exp1_2}
\end{figure*}

\noindent\textbf{Masked-guided Blending Adapter.} Despite spatial pixel-level conditions have been given, we observe significant distortion artifacts in the background, resulting in a low fidelity of the synthesized video. Therefore, we introduce a \textit{Masked-guided Blending Adapter} $\boldsymbol{\mathcal{F}}_{MB}$ into VAE Decoder $\boldsymbol{\mathcal{D}}$, forming $\boldsymbol{\mathcal{D}}^{'}$, which utilizes a mask to directly copy the masked region of the $k$-th decoder feature $\boldsymbol{F}^{k}_{vd}$ while combining the unmasked region of the $k$-th background feature $\boldsymbol{F}^{k}_{ve}$ from the VAE encoder. The dilated mask $\boldsymbol{m}$ and background $\boldsymbol{I}_{bg}$ are produced by MODNet~\cite{ke2022modnet} $\boldsymbol{\mathcal{G}}_{MOD}$ and Inpaint-Anything~\cite{yu2023inpaint} $\boldsymbol{\mathcal{G}}_{IA}$ respectively. The blending process is formulated as:
\begin{equation}
    \begin{aligned}
    \hat{\boldsymbol{F}}_{vd}^k = \boldsymbol{F}_{vd}^k \otimes \boldsymbol{m} + \text{Conv}(\boldsymbol{F}_{ve}^k) \otimes \tilde{\boldsymbol{m}},
    \end{aligned}
\end{equation}
where $\otimes$ is element-wise multiplication, $\tilde{\boldsymbol{m}}=1-\boldsymbol{m}$. Experimentally, we inset this module into the decoder layer with the resolution of 512. The training only involves a frozen autoencoder and a trainable convolution layer, under the supervision of reconstruction and VGG losses. With this design, the synthesized video maintains a consistent background, and the fusion edge is seamlessly harmonious. Note that the above three adapters are trained sequentially.

\noindent\textbf{Autoregressive Inference.} The designed CCF has diverse and controllable image generation capabilities. We incorporate an \textit{Autoregressive Inference} to further generate diverse and temporally coherent videos. As shown in Alg.~\ref{alg:ai}, unlike DiffTalk~\cite{shen2023difftalk} starting from a given frame, for the first frame, we switch CCF to diverse mode, \ie, only receive audio-enhanced CLIP embedding $\boldsymbol{Y}$ and encoded 3D face mesh features $\boldsymbol{\mathcal{E}}(\boldsymbol{I}_{rd}^1)$, generating faces $\hat{\boldsymbol{I}_1}$ with varying appearance while maintaining the consistent geometry with the audio. For the subsequent frames, we further apply three adapters to achieve controllable generation. Specifically, $\boldsymbol{I}_{id}^{i}$ and $\boldsymbol{I}_{bg}^{i}$ are $\hat{\boldsymbol{I}}_1$ and $\boldsymbol{\mathcal{G}}_{IA}(\hat{\boldsymbol{I}}_1)$ all the time, $\boldsymbol{I}_{rd}^{i}$ is the mesh at time stamp $i$, and $\boldsymbol{I}_{ad}^{i}$ is the masked former frame $\hat{\boldsymbol{I}}_{i-1}$, $\boldsymbol{m}_{i}=\boldsymbol{\mathcal{G}}_{MOD}(\hat{\boldsymbol{I}}_{i-1})$ since the difference between adjacent frames is small.

\section{Experiments}
\label{sec:experiments}

\subsection{Experimental Setup}
\noindent\textbf{Datasets.} 
We adopt three talking face dataset MEAD~\cite{wang2020mead}, VoxCeleb2~\cite{chung2018voxceleb2}, and HDTF~\cite{zhang2021flow} in our experiments.

\noindent\textbf{Metrics.} We adopt PSNR, SSIM, and FID~\cite{heusel2017gans} to measure the image quality of synthesized videos. The distance between the landmarks of the mouth (LMD) and confidence score (Sync) proposed in SyncNet~\cite{chung2017out} is used to measure the audio-visual synchronization. Besides, we compute emotion accuracy (EACC) to measure the generated emotions. Furthermore, we compute the cosine similarity of geometry (Shape) and semantics (Age, Gender, Emotion), and overall quality (FID) between the ground truth and generated faces inferred from audio by using several off-the-shelf tools, D3DFR~\cite{deng2019accurate}, FairFace~\cite{karkkainenfairface}, and FAN~\cite{meng2019frame}.

\noindent\textbf{Implementation Details.} 
For PAD training, we adopt VoxCeleb2 for identity disentanglement and MEAD training set for content and emotion. The audios are pre-processed to 16kHz and converted to mel-spectrogram with 80 Mel filter-bank, as Wav2Lip. The length of the sampled video clip is $L=32$. For CCF training, the input videos are cropped and resized to $512 \times 512$. We adopt MEAD and HDTF as training sets and optimize TIA, SCA, and MBA sequentially. It takes about 2.5 days totally when trained on 8 V100 GPUs. $N=8$ and $M$ is the length of the prompts in SCA. During quantitative assessment, we sample 12 identities from MEAD and 10 from HDTF, which serve as the test set. Denoising step $T$ is set as 50 for both training and inference.

\subsection{Comparison with State-of-the-Art Methods.}

\noindent\textbf{Qualitative Results.} 
In this section, we first perform audio-to-face method comparisons with CMP~\cite{wu2022cross}. As shown in Fig.~\ref{fig:exp1_1}, we display the generated 3D meshes and their corresponding faces of our method in rows 4, 5, and CMP in rows 6, 7. The real faces in the first row provide geometry and semantics references. It can be seen that CMP fails to generate desired shapes and the synthesized faces have blurred details with obvious artifacts. By contrast, benefiting from progressive disentanglement and LDMs, our method produces more accurate geometry, including shape, lip movement, emotion styles, and high-quality realistic textures. Besides, in each sample, the first two columns correspond to the same audio, while the third represents a different audio of the same person. Across different audio clips, our method produces more consistent results than CMP. We further compare our method with talking face methods including Wav2Lip, PC-AVS, EAMM, and SadTalker, which are reproduced from their officially released codes. As shown in Fig.~\ref{fig:exp1_2}, in terms of emotion accuracy, audio-visual synchronization, and image quality, our method outperforms these SOTA methods. Notably, for the third sample, other competitors adhere the two stage paradigm, \ie, Audio-to-Image and then Image-to-Video, while our method autoregressively infers video from the given audio. 

\begin{figure}[t!]
	\centering
	\includegraphics[width=0.44\textwidth]{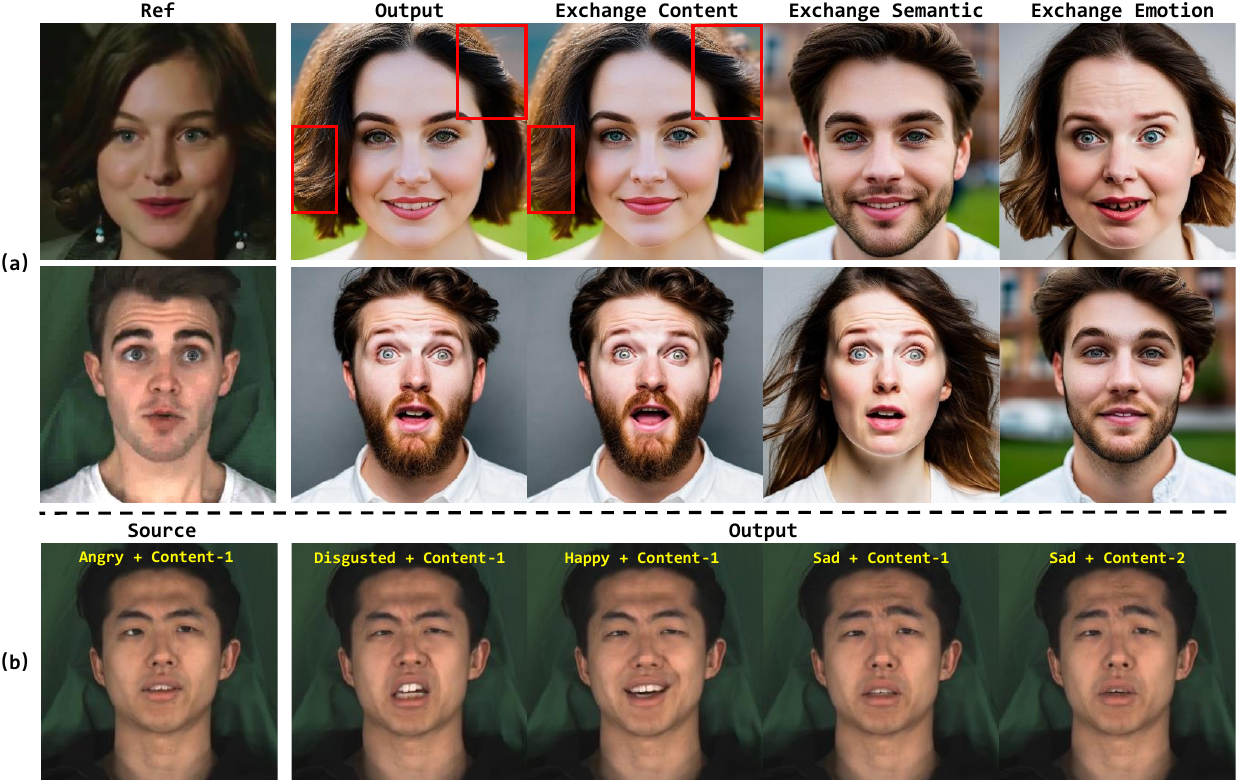}
	\caption{Illustration of disentangled controllability. (a) is under the diverse mode and (b) is under the coherent mode.
	}
	\label{fig:exp2_1}
\vspace{-0.5em}
\end{figure}

\begin{table}[t]
   \centering
   \scriptsize
   \renewcommand\arraystretch{1.0}
   \setlength\tabcolsep{4pt}
   \begin{tabular}{C{30pt}C{25pt}C{25pt}C{29pt}C{31pt}C{25pt}}
      \toprule
      Method & Shape $\uparrow$ & Age $\uparrow$ & Gender $\uparrow$  & Emotion $\uparrow$ & FID $\downarrow$ \\
      \midrule
      CMP~\cite{wu2022cross} & 0.60           & 0.41      & 0.87     & 0.23  &    228.05   \\
      Ours & \textbf{0.75}  & \textbf{0.58}   & \textbf{0.93} & \textbf{0.76} & \textbf{25.78}              \\
      \bottomrule
   \end{tabular}
   \caption{Quantitative comparison with CMP. The value is the average of all samples on two test sets (same as the Tab.~\ref{tab:sota_2}).}
   \label{tab:sota_1}
\vspace{-0.5em}
\end{table}

\begin{table}[t]
   \centering
   \scriptsize
   \renewcommand\arraystretch{1.0}
   \setlength\tabcolsep{4pt}
   \begin{tabular}{C{40pt}C{25pt}C{22pt}C{24pt}C{22pt}C{24pt}C{24pt}}
      \toprule
      Method & EACC $\downarrow$ & LMD $\downarrow$ & Sync $\uparrow$  & FID $\downarrow$ & PSNR $\uparrow$ & SSIM $\uparrow$ \\
      \midrule
      Wav2Lip~\cite{prajwal2020lip} & 0.129  & 2.96   & \textbf{5.93} 	     & 40.79             & 29.05             & 0.59              \\
      PC-AVS~\cite{zhou2021pose} & 0.102   & \underline{2.71}   & 5.75	      &  45.84             & 29.28             & 0.59             \\
      EAMM~\cite{ji2022eamm} & \underline{0.097}   & 2.75   & 5.46           & 50.55              & 29.14             & 0.61              \\
      SadTalker~\cite{zhang2023sadtalker} & 0.121   & 2.80   & 5.84  	   & 31.62              & 29.55             & \underline{0.64}              \\
      \midrule
      Ours-C & 0.113  & 2.82    & 5.86       & \textbf{28.09}              & \underline{29.59}             & 0.60  \\
      Ours-E & \textbf{0.090}  & \textbf{2.67}   & \underline{5.91}   & \underline{28.37}  & \textbf{29.66} & \textbf{0.67}              \\
      \bottomrule
   \end{tabular}
   \caption{Quantitative comparison with state-of-the-art talking face methods on two test sets.}
   \label{tab:sota_2}
\vspace{-0.5em}
\end{table}

\noindent\textbf{Quantitative Results.} As shown in Tab.~\ref{tab:sota_1}, our method excels over CMP across all metrics, especially in FID, which is consistent with the qualitative results in Fig.~\ref{fig:exp1_1}. Moreover, we observe that gender is relatively easy to predict, while others are more challenging. Yet, our method still gains significant improvements. We further present the quantitative comparison in Tab.~\ref{tab:sota_2} of talking face generation, which reveals our method can achieve the best performance in most metrics. Wav2Lip surpasses all the competitors on Sync due to the supervision of the synchronization scoring model. We further report the user study in supplementary materials.

\subsection{Further Analysis}

\begin{figure}[t!]
	\centering
	\includegraphics[width=0.43\textwidth]{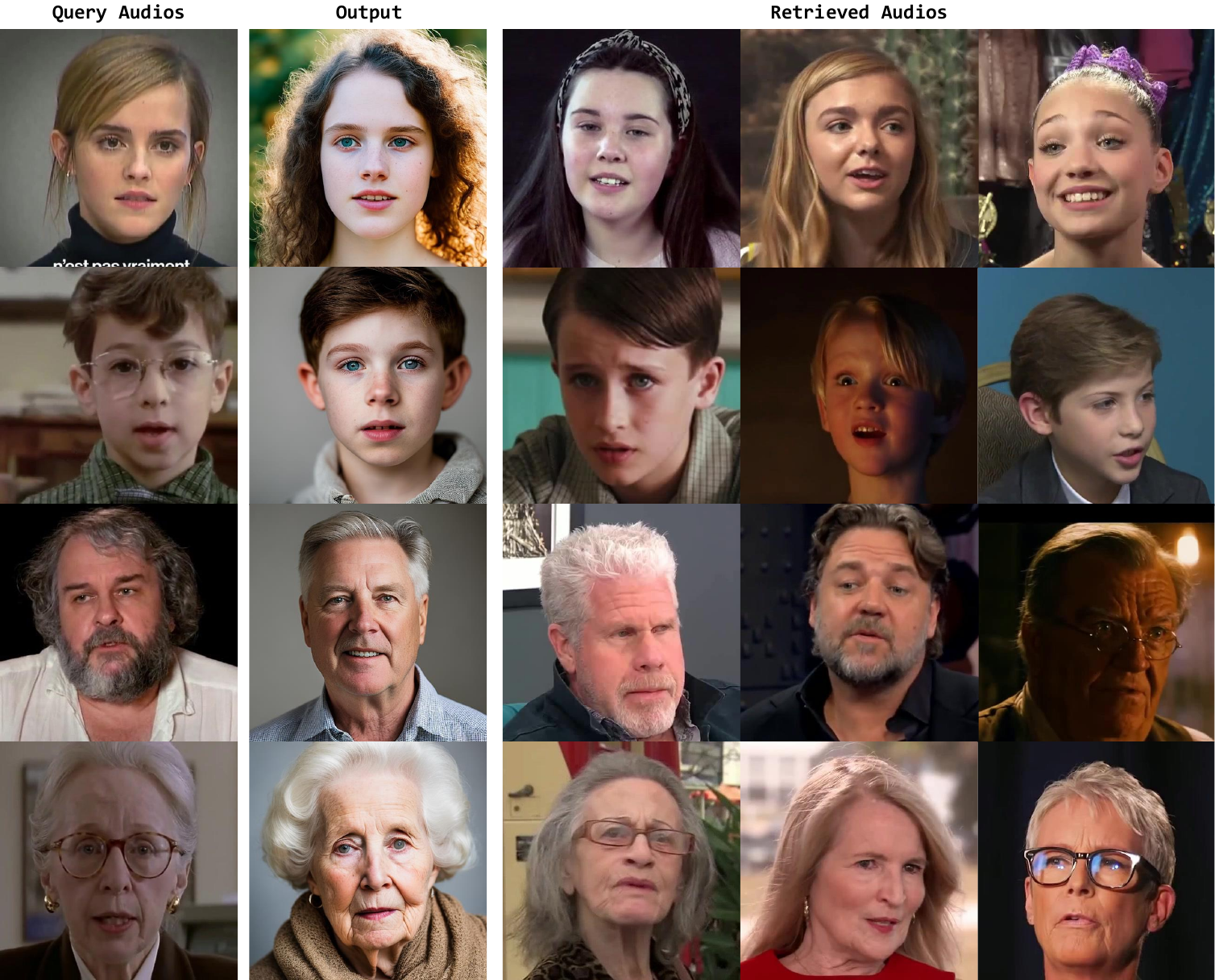}
	\caption{Illustration of the image retrieval using the identity semantic features. The faces (column 1) only shown for reference.
	}
	\label{fig:exp2_2}
\vspace{-0.5em}
\end{figure}

\begin{figure}[t!]
	\centering
	\includegraphics[width=0.37\textwidth]{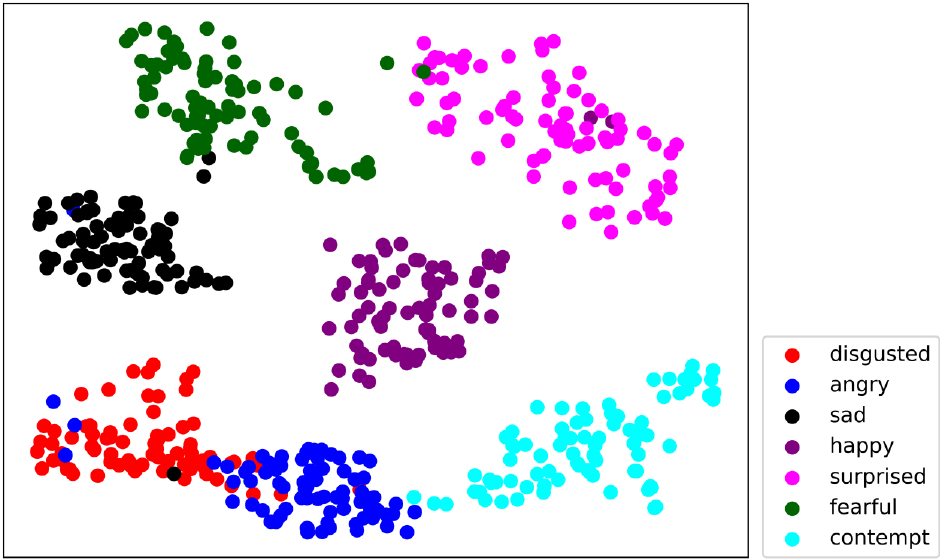}
	\caption{Clusters of the disentangled emotion embeddings. Different emotions are clustered in different regions.
	}
	\label{fig:exp2_3}
\vspace{-0.5em}
\end{figure}

\noindent\textbf{Disentangled Controllability.} To demonstrate the facial factor disentanglement of our method, we first sample two distinctly different faces and exchange one factor at a time in diverse mode. Although our method independently controls the desired factor, \ie, content (lip movement), semantics (gender and age), and emotion styles in columns 3, 4, 5 of Fig.~\ref{fig:exp2_1}(a), the results inherit the weaknesses of the LDM in preserving visual consistency. As the red rectangles marked in row 1, despite only editing the mouth shape, they exhibit noticeable difference in appearance, which indicates that simply generating frames and splicing them together cannot create visually and temporally consistent videos. Furthermore, we conduct a more rigorous analysis in coherent mode. The emotion control (cols. 1-4) and content editing (cols. 4-5) are illustrated in Fig.~\ref{fig:exp2_1}(b), verifying the effectiveness of the disentanglement.

\noindent\textbf{Interpretability of Disentanglement.} We further perform qualitative experiments to demonstrate the disentanglement of learned identity and emotion representation. Firstly, we randomly sample four audios of different ages and genders from the CelebV-HQ~\cite{zhu2022celebv} dataset and select their most consistent pairs according to the cosine similarity between two semantic features. We display the corresponding images for visual comparisons in Fig.~\ref{fig:exp2_2}, where the retrieved audios have similar ages and genders to those of the query. We also attach the generated face in column 2, which closely resembles the conditions in column 1, illustrating the effectiveness of our method. Secondly, to evaluate the disentanglement of the emotion, we use t-SNE~\cite{van2008visualizing} to visualize the emotion latent space in Fig.~\ref{fig:exp2_3}. It is obvious that the different emotions are clustered into separate groups. Based on these analyses, we conclude that the PAD contributes to the decoupling entangled facial cues from the audio.

\subsection{Ablation Study and Efficiency Evaluation}

\begin{figure}[t!]
    \centering
	\begin{minipage}{0.98\linewidth}
		\centering
		\setlength{\abovecaptionskip}{0.28cm}
		\includegraphics[width=\linewidth]{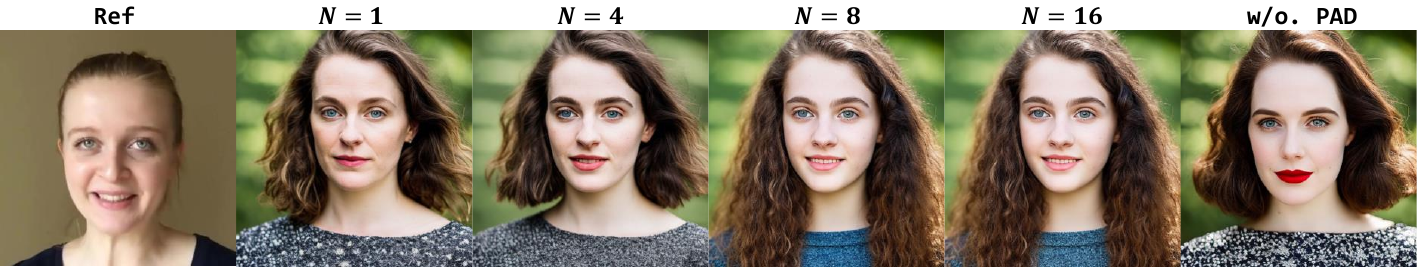}
		\label{fig:exp3_1}
	\end{minipage}
	\hfill
	\begin{minipage}{0.98\linewidth}
		\centering
        \scriptsize
        \renewcommand\arraystretch{1.0}
        \setlength\tabcolsep{4pt}
        \vspace{-0.3cm}
        \begin{tabular}{C{20pt}C{32pt}C{32pt}C{32pt}C{32pt}C{32pt}}
          \toprule
           & N=1 & N=4 & N=8 & N=16 & w/o. PAD  \\
          \midrule
          Age &  0.23 & 0.42 & \textbf{0.51} & \underline{0.50} &0.38 \\
          Gender &  0.90 & 0.91 & \underline{0.91} & \textbf{0.92} &0.91 \\
          Emotion &  0.49 & 0.62 & \underline{0.71} & \textbf{0.73} & 0.53 \\
        \bottomrule
      \end{tabular}
        \caption{Ablation study on the number of pseudo-word tokens in TIA. We do not apply SCA and MBA in this stage.}
        \label{fig:exp3_1}
	    \end{minipage}
\vspace{-0.5em}
\end{figure}

\begin{figure}[t!]
	\centering
	\includegraphics[width=0.46\textwidth]{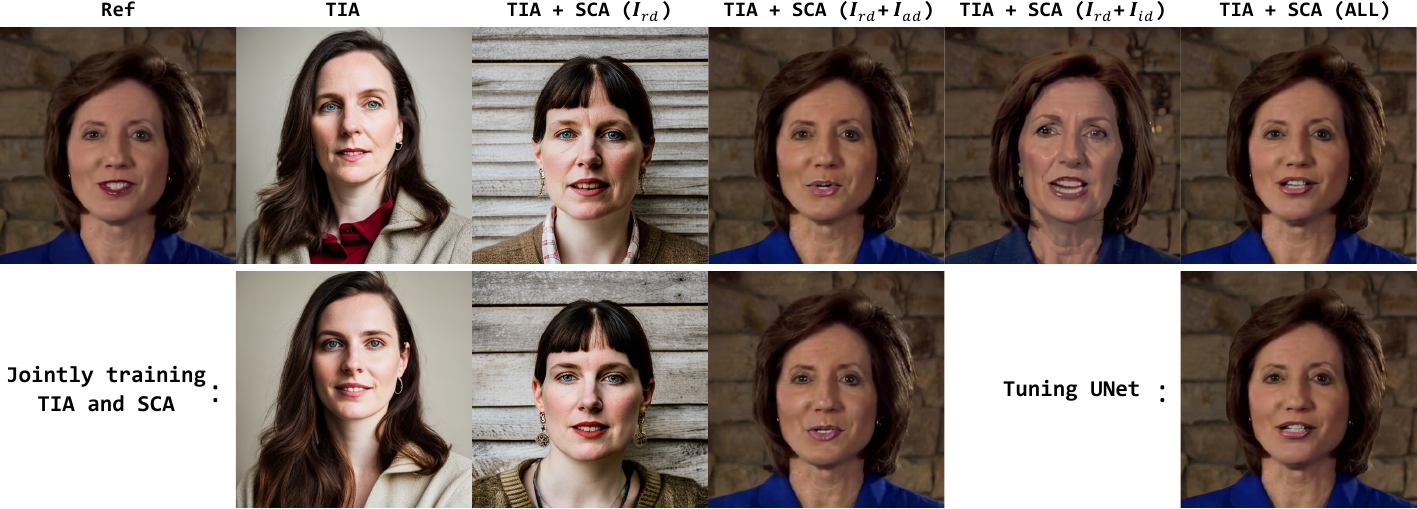}
	\caption{Ablation study on the design of the conditions in SCA.
	}
	\label{fig:exp3_2}
  \vspace{-0.5em}
\end{figure}

\noindent\textbf{Impact of Disentangled Module in PAD.} To verify the superiority of PAD, we show the generated meshes under different stages in Fig.~\ref{fig:exp1_1} rows 2-4. The first stage handles identity disentanglement, and the corresponding meshes accurately reflect desired face shapes and outlines. The following content disentanglement ensures that the mouth movements synchronize with the spoken phonemes. The final emotion disentanglement enhances the facial global cues, leading to highly consistent meshes with the reference faces, \eg, more accurate lips with fearful emotion (the right sample). Note that the left sample with neutral shows no noticeable changes. Besides, the last two rows of Tab.~\ref{tab:sota_2} can also prove the effectiveness of PAD.

\noindent\textbf{Impact of Pseudo-word Tokens in TIA.} In Fig.~\ref{fig:exp3_1}, we show the results when varying the number of pseudo-word tokens in TIA. We observe that too few tokens can not synthesize semantically consistent results, while too many do not lead to further visual improvement. To balance the computational load and performance, we set $N$ as 8 experimentally. Besides, 
the sixth column indicates that when directly inputting audio features into TIA, the output suffers degradation due to insufficient decoupling. We further attach the quantitative results in Fig.~\ref{fig:exp3_1}. Since TIA does not involves face geometry, shape metric is not displayed. These results are based on the 100 faces sampled from CelebV-HQ.

\noindent\textbf{Impact of Conditions in SCA.} The conditions in SCA are the critical design in this work for generating controllable and coherent frames. As shown in Fig.~\ref{fig:exp3_2} row 1, the model with TIA can generate semantically consistent faces but fails to preserve the geometrical information (column 2). Introducing the SCA module with $\boldsymbol{I}_{rd}$ input tackles the spatial misalignment (column 3). The adjacent frame $\boldsymbol{I}_{ad}$ is further incorporated trying to maintain the appearance like the reference face (column 4), but it introduces obvious artifacts, especially in the mouth area. When taking $\boldsymbol{I}_{rd}$ and $\boldsymbol{I}_{id}$ as input, as previously mentioned, the frozen LDMs encounter difficulties in learning complex transformations, resulting in both the appearance and geometry are not well aligned (column 5). In this work, we employ all three inputs together for SCA (column 6), where $\boldsymbol{I}_{rd}$ is responsible for the facial structure, $\boldsymbol{I}_{ad}$ for the appearance, while $\boldsymbol{I}_{id}$ further refines the details and enhances the image quality.

\begin{figure}[t!]
	\centering
	\includegraphics[width=0.46\textwidth]{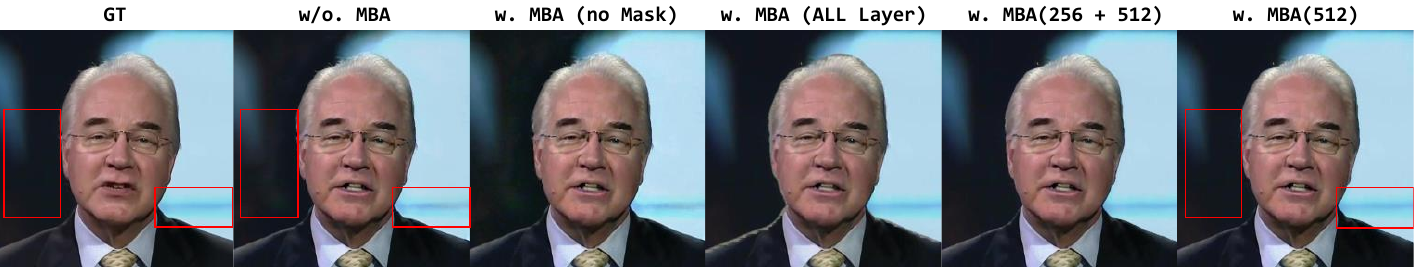}
	\caption{Ablation study on the design of the blending structure in MBA. This case is sampled from Fig.~\ref{fig:exp1_2}.
	}
	\label{fig:exp3_3}
 \vspace{-0.5em}
\end{figure}

\noindent\textbf{Impact of Blending Structure in MBA.} In Fig.~\ref{fig:exp3_3}, the result in the second column is produced by the method without MBA, which fails to preserve the background details. To solve this challenge, we design several variants to explore the blending structure. Without mask guidance, the synthesized background still differs from the ground truth (column 3). For this reason, we introduce the mask into all VAE decoder layers (column 4). Due to the edge aliasing in the low-resolution mask, although it solves the background issue, obvious artifacts appear around the blending edges. When applied at 512 resolution (column 6), the generated face achieves the best visual performance, achieving consistent background and seamless edge blending.

\noindent\textbf{Impact of Training Strategy in CCF.} We explore the effectiveness of sequential training in CCF. As shown in Fig.~\ref{fig:exp3_2}, when jointly training TIA and SCA, the former does not work well as facial semantics can take shortcuts, \ie, learning from $\boldsymbol{I}_{id}$ and $\boldsymbol{I}_{ad}$ instead of token embeddings $\boldsymbol{Y}$.

\noindent\textbf{Efficiency Evaluation.} We follow the DiffTalk to open the UNet parameters during SCA training, the number of trainable parameters increases from 282.9M to 1.15G. Furthermore, tuning the UNet does not bring significant benefits in Fig.~\ref{fig:exp3_2}. Thus, our method is effective and user-friendly.

\section{Conclusions}
\label{sec:conclusion}

We propose a novel framework to simulate the process of Listening and Imagining. Firstly, PAD simplifies the decoupling process and improves the synchronization of predicted representations with audio. Secondly, CCF further generates realistic and diverse talking faces. Extensive experiments demonstrate the effectiveness of our approach. 
{
    \small
    \bibliographystyle{ieeenat_fullname}
    \bibliography{main}
}


\end{document}